\documentclass[sigconf]{acmart}
\AtBeginDocument{%
  }

\copyrightyear{2026}
\acmYear{2026}
\setcopyright{cc}
\setcctype{by}
\acmDOI{10.1145/3767308.3835028}
\acmConference[MM '26]{Proceedings of the 34th ACM International Conference on Multimedia}{November 10--14, 2026}{Rio de Janeiro, Brazil}
\acmBooktitle{Proceedings of the 34th ACM International Conference on Multimedia (MM '26), November 10--14, 2026, Rio de Janeiro, Brazil}
\acmISBN{979-8-4007-2213-4/2026/11}

\usepackage{balance}
\usepackage{color,xcolor}
\usepackage{colortbl}
\newcommand{\pub}[1]{\color{gray}{\scriptsize{[{#1}]}}}
\usepackage{multirow}
\usepackage{bm}
\usepackage{graphicx}
\usepackage{array}
\usepackage[table]{xcolor}
\usepackage{amsmath}
\usepackage{makecell}
\usepackage{arydshln}
\usepackage{hhline}
\usepackage{bbm}

\begin{document}

\title{Acoustically Grounded Cost Learning for Open-Vocabulary Audio-Visual Semantic Segmentation}

\author{Tianrui Hui}
\affiliation{%
  \institution{Hefei University of Technology}
  \city{Hefei}
  \state{Anhui}
  \country{China}
}

\author{Shaofei Huang}
\authornote{Corresponding author.}
\affiliation{%
  \institution{Hefei University of Technology}
  \city{Hefei}
  \state{Anhui}
  \country{China}
}

\author{Qisong Han}
\affiliation{%
  \institution{Cardiff University}
  \city{Cardiff}
  \state{Wales}
  \country{United Kingdom}
}

\author{Yaxiong Wang}
\affiliation{%
  \institution{Hefei University of Technology}
  \city{Hefei}
  \state{Anhui}
  \country{China}
}

\author{Lechao Cheng}
\affiliation{%
  \institution{Hefei University of Technology}
  \city{Hefei}
  \state{Anhui}
  \country{China}
}

\author{Zhedong Zheng}
\affiliation{%
  \institution{University of Macau}
  \city{Macau}
  \country{China}
}

\author{Zhun Zhong}
\affiliation{%
  \institution{Hefei University of Technology}
  \city{Hefei}
  \state{Anhui}
  \country{China}
}

\author{Richang Hong}
\affiliation{%
  \institution{Hefei University of Technology}
  \city{Hefei}
  \state{Anhui}
  \country{China}
}

\author{Meng Wang}
\affiliation{%
  \institution{Hefei University of Technology}
  \city{Hefei}
  \state{Anhui}
  \country{China}
}

\renewcommand{\shortauthors}{Tianrui Hui et al.}

\begin{abstract}
  Open-Vocabulary Audio-Visual Semantic Segmentation (OV-AVSS) aims to perform pixel-level segmentation of sound-emitting objects from an open set of categories. The previous method relies on a class-agnostic foreground definition, which groups semantically diverse objects into a heterogeneous positive set, causing the model to learn unstable sounding patterns and produce unreliable proposals. To address this, we reformulate the objective to be category-specific and propose a novel Acoustically Grounded Cost Learning (AGCL) framework to transform the static, audio-agnostic visual-text priors into dynamic, audio-grounded cost representations. For intra-category soundingness discovery, we devise Audio-Modulated Cost Generation (AMCG) and Audio-Guided Temporal Aggregation (AGTA) modules to enable both frame-level sounding region highlighting and video-level temporal refinement with a low-intrusive audio injection mechanism. For inter-category distractor discrimination, we introduce a Synergistic Distractor Mining (SDM) strategy, which selectively penalizes acoustically and semantically confusing negative categories to learn more discriminative decision boundaries. Extensive experiments on the AVSBench-OV dataset demonstrate that our method significantly outperforms previous state-of-the-art approaches, particularly on unseen categories. Code is available at \url{https://github.com/spyflying/AGCL}.
\end{abstract}

\ccsdesc[500]{Computing methodologies~Video segmentation}
\ccsdesc[500]{Computing methodologies~Scene understanding}

\keywords{Audio-Visual Semantic Segmentation, Open-Vocabulary, Cost Volume Learning}


\maketitle

\section{Introduction}
\label{sec:intro}
The goal of the Open-Vocabulary Audio-Visual Semantic Segmentation (OV-AVSS)~\cite{guo2024open} task is to achieve pixel-level segmentation and semantic recognition of sound-emitting objects from a given video and its associated audio across an open set of semantic categories.
Unlike the conventional AVSS task~\cite{zhou2022audio,zhou2025audio} that operates under a closed-set assumption, OV-AVSS requires the model to generalize beyond training categories and correctly identify both seen and unseen ones during inference, making this task substantially more challenging.
Owing to this demand for generalization to unseen and unheard categories, OV-AVSS holds great potential for real-world applications such as robotic perception~\cite{zhang2025moma,he2024progressive}, intelligent content creation~\cite{di2021video,he2024customize}, embodied spatial reasoning~\cite{ju2026instruction,wu2024masked}, etc.

\begin{figure}[!t]
   \centering
   \includegraphics[width=\linewidth]{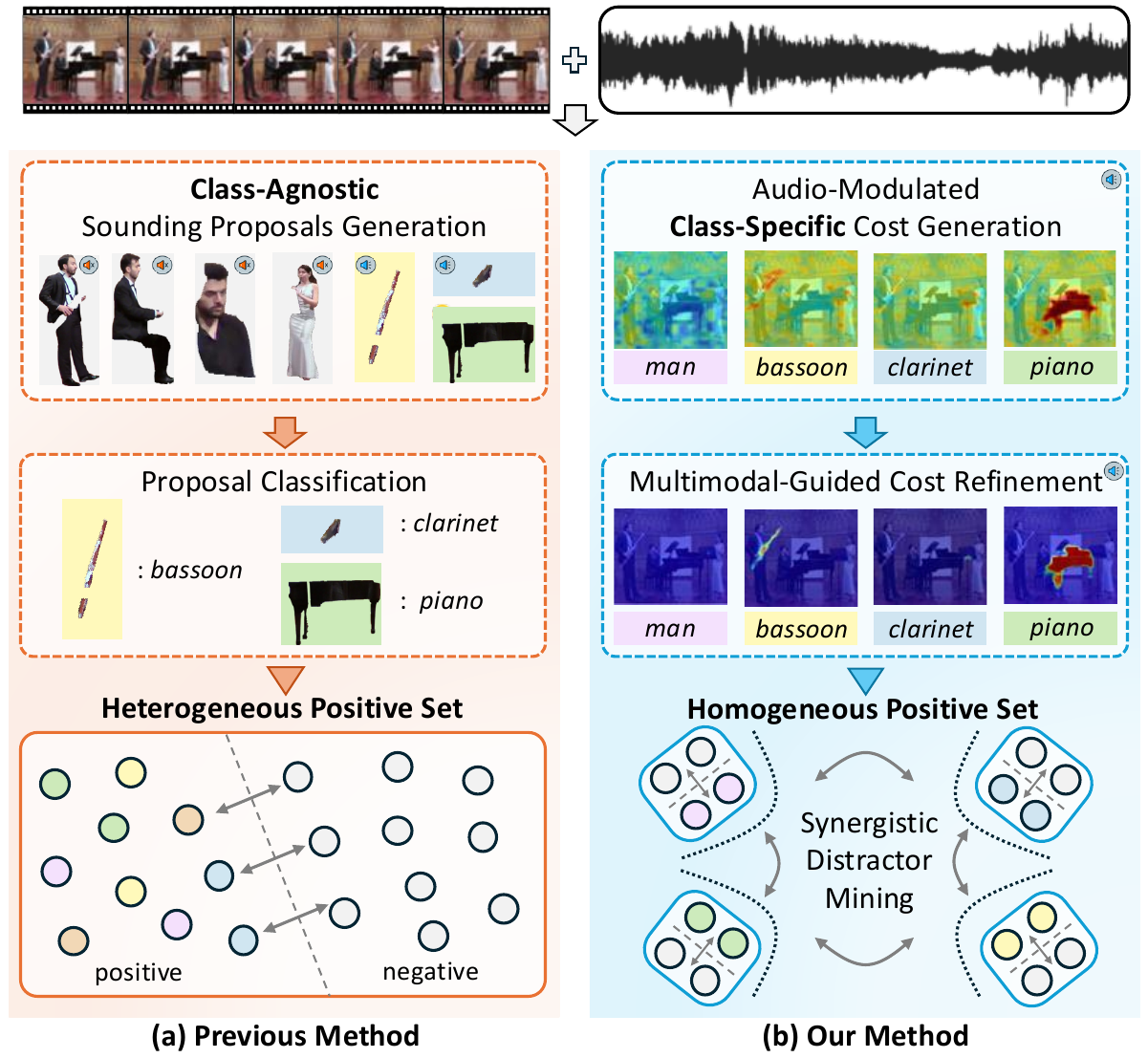}
   \caption{Motivation of our method. (a) The previous method adopts a class-agnostic foreground definition, grouping semantically diverse objects into a highly heterogeneous positive set. (b) Our method reformulates the foreground definition of OV-AVSS to be class-specific. By explicitly grounding acoustic events to cost representations, we achieve accurate intra-category soundingness discovery and robust inter-category distractor discrimination.}
   \label{fig:intro}
\end{figure}

As shown in Figure~\ref{fig:intro}(a), the previous method~\cite{guo2024open} first generates class-agnostic sounding mask proposals via an audio-augmented segmentation model, then classifies them by offline feature matching with category texts using a frozen CLIP~\cite{radford2021learning} model.
However, foreground is defined event-wise as ``\textit{\textbf{any object} that emits the given sound}'' in the proposal generation stage, regardless of its semantic category.
Such a category-agnostic definition groups semantically diverse objects under the same positive label, yielding a highly heterogeneous positive set with poor inner consistency.
Consequently, the model tends to fit unstable soundingness rather than category-consistent objectness, resulting in unreliable proposals.
These low-quality proposals further propagate errors to the proposal classification stage, degrading the performance.
To address this issue, we reformulate the foreground definition in OV-AVSS as ``\textit{\textbf{object of a specific category} that emits the given sound}'', turning the original category-agnostic objective into a category-specific one.
To accommodate the reformulated objective, we adopt the cost learning paradigm~\cite{cho2024cat}, where CLIP-derived visual-text matching costs naturally provide localization priors for individual categories.
However, such priors only capture spatial existence rather than sounding dynamics for each category, yielding two primary challenges for the reformulation: \textbf{intra-category soundingness discovery}, \textit{i.e.}, isolating objects ``\textit{emitting sound}'' from silent ones within each specific category, and \textbf{inter-category distractor discrimination}, \textit{i.e.}, distinguishing the category of ``\textit{given sound}'' from acoustically confusing ones whose sounding characteristics are inadequately captured by visual-text priors.
Therefore, we propose an Acoustically Grounded Cost Learning (AGCL) framework to transform the static, audio-agnostic visual-text priors into dynamic, audio-grounded cost representations.

For the first challenge, the AGCL framework features a low-intrusive audio injection mechanism that meticulously incorporates acoustic cues during both the generation and refinement of the cost maps, thereby highlighting sounding objects within each category while preserving the original open-vocabulary recognition capability.
In the cost generation phase, we devise an Audio-Modulated Cost Generation (AMCG) module, which modulates frame-wise visual features in the CLIP vision encoder with audio cues before visual-text correlation.
This modulation explicitly activates sounding regions on the cost map of each category without destructively shifting the original visual feature distributions.
In the cost refinement phase, we further devise an Audio-Guided Temporal Aggregation (AGTA) module, which refines category-wise cost representations across the video by aggregating cost embeddings according to audio-derived temporal similarity, thereby improving the temporal consistency of soundingness discovery.
Together, these two modules enable both frame-level sounding region highlighting and video-level temporal refinement, leading to more reliable sounding object discovery within each category.

To tackle the second challenge, we introduce the Synergistic Distractor Mining (SDM) strategy for robust inter-category discrimination.
Since visual-text priors inherently lack sounding characteristics, standard cost learning struggles to establish clear decision boundaries among acoustically confusing categories.
To overcome this limitation, rather than uniformly suppressing all negative categories during training, our SDM leverages external audio-language knowledge to selectively penalize the most confusing distractors based on comprehensive acoustic and semantic cues.
Specifically, acoustic distractors are identified by the similarity between the input audio and category texts in the embedding space of a pretrained audio-language model, while semantic distractors are identified by the textual feature similarity between ground-truth and other categories.
By explicitly penalizing these highly confusing distractors, SDM effectively sharpens the decision boundaries among easily confused categories, thereby encouraging better discrimination of unseen categories.

Our contributions are summarized as follows:
(1) We reformulate the foreground definition in OV-AVSS to be class-specific and propose a novel Acoustically Grounded Cost Learning (AGCL) framework to transform the static, audio-agnostic visual-text priors into dynamic, audio-grounded cost representations.
(2) For intra-category soundingness discovery, AGCL includes Audio-Modulated Cost Generation and Audio-Guided Temporal Aggregation modules to highlight sounding regions in each category while preserving the open-vocabulary capabilities.
For inter-category distractor discrimination, we introduce the Synergistic Distractor Mining strategy to penalize semantically and acoustically confusing distractors for better generalization.
(3) Extensive experiments on the AVSBench-OV dataset verify that our method outperforms previous state-of-the-art methods by large margins, especially for unseen categories (mIoU $29.14\%$ \textit{vs.} $45.59\%$).

\begin{figure*}[t]
   \centering
   \includegraphics[width=1.0\textwidth]{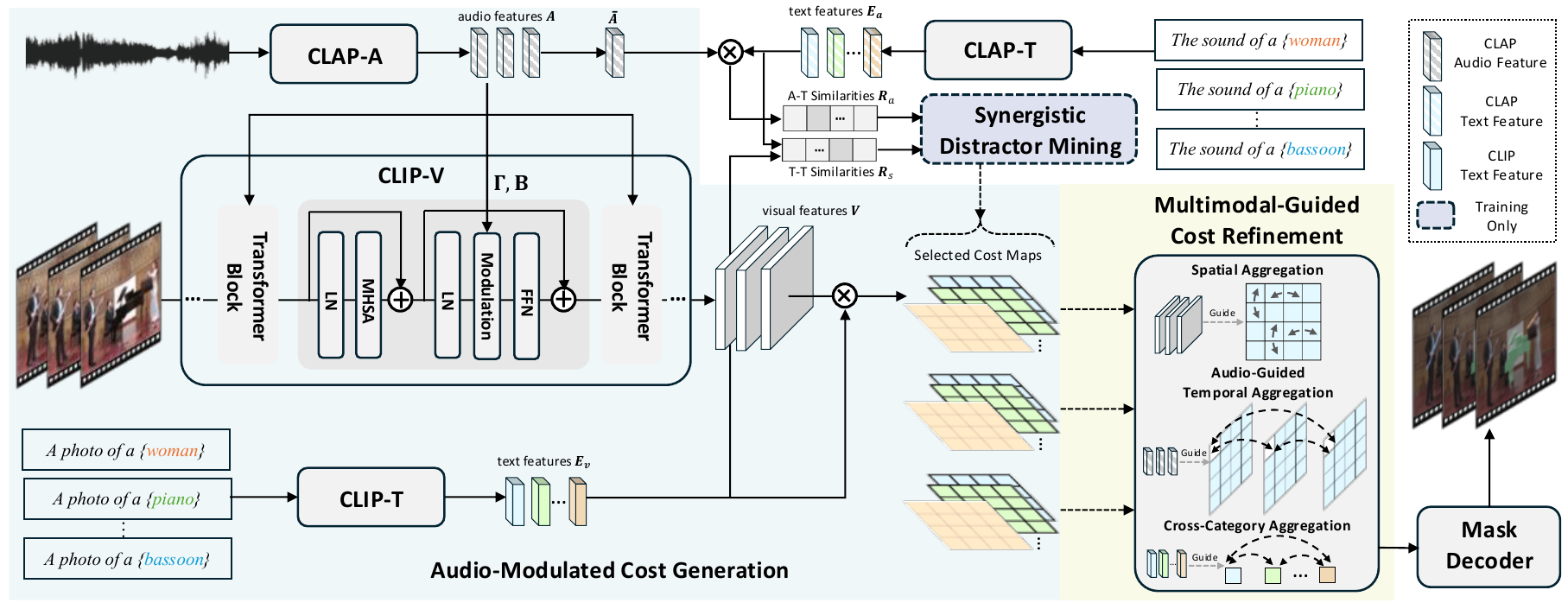}
   \caption{Overall architecture of our method. In the Audio-Modulated Cost Generation module, category-wise cost maps are obtained by correlating visual features from CLIP-V with category text features from CLIP-T. Audio features from CLAP-A are injected into CLIP-V to modulate visual features and softly highlight sounding regions. The resulting cost maps are further refined in the Multimodal-Guided Cost Refinement module to produce sharper spatial boundaries and better temporal consistency. The refined cost maps are then fed into a mask decoder to generate category-wise sounding object segmentation masks. During training, we apply the Synergistic Distractor Mining (SDM) strategy to the generated cost maps to select acoustically and semantically confusing distractor categories for more discriminative learning.}
   \label{fig:pipeline}
\end{figure*}

\section{Related Work}
\label{sec:related_work}
\subsection{Audio-Visual Segmentation}
Zhou \textit{et al.}~\cite{zhou2022audio,zhou2025audio} first introduce the Audio-Visual Segmentation (AVS) task and collect two datasets, namely AVSBench-Object and AVSBench-Semantic.
The former focuses on binary foreground-background segmentation of sounding objects, while the latter further requires identifying the semantic categories of the sounding objects in addition to segmentation.
Many subsequent studies~\cite{huang2023discovering,li2023catr,liu2023audio,wang2024prompting,sun2024unveiling,gao2024avsegformer,chen2024unraveling,yang2024cooperation,li2024selm,wang2024ref,huang2025unleashing,huang2025revisiting} adopt the audio-query design and achieve consistent performance gains.
COMBO~\cite{yang2024cooperation} introduces a bilateral entanglement mechanism at the pixel, modality, and temporal levels for more effective audio-visual fusion.
VCT~\cite{huang2025revisiting} constructs queries in a vision-centric manner, alleviating the perception ambiguity of audio queries and the loss of visual details.
Beyond architectural evolution, UFE~\cite{liu2024audio} exploits motion and semantic cues from unlabeled neighboring and distant frames to improve performance, while ICF~\cite{zha2025implicit} introduces an implicit counterfactual framework that balances audio-visual modalities through implicit text bridging and counterfactual learning.
This paper focuses on the OV-AVSS~\cite{guo2024open} task, which extends the closed-set semantic categories in AVS to an open-vocabulary scenario.
Different from the previous method~\cite{guo2024open} that relies on a class-agnostic foreground definition, we reformulate the task to model category-specific sounding situations.
By explicitly grounding acoustic events to cost representations, our AGCL framework tackles the challenges of intra-category soundingness discovery and inter-category distractor discrimination for more effective open-vocabulary segmentation.

\subsection{Open-Vocabulary Semantic Segmentation}
Open-Vocabulary Semantic Segmentation (OVSS) aims to enable models to perform pixel-level segmentation for arbitrary semantic categories, including those unseen during training.
Current OVSS methods can generally be categorized into two paradigms: two-stage pipelines~\cite{ding2022decoupling,ding2022open,huynh2022open,liang2023open,xu2022simple,liu2024open,li2025mask} and end-to-end pipelines~\cite{xu2022groupvit,xu2023open,xu2023side,wang2023transferring,cho2024cat,xie2024sed,zhao2025dpseg}.
Two-stage pipelines typically first extract object proposals and then classify them using CLIP~\cite{radford2021learning}.
For example, SCAN~\cite{liu2024open} enhances the alignment between proposals and category texts by incorporating CLIP-based semantic priors and contextual calibration.
Mask-Adapter~\cite{li2025mask} extracts semantic activation maps from masks and enforces consistency to better align them with CLIP features.
Among end-to-end pipelines, SAN~\cite{xu2023side} designs side-adapters on top of a frozen CLIP to predict masks and attention biases, achieving efficient OVSS.
CAT-Seg~\cite{cho2024cat} constructs and aggregates a cost volume between visual and textual features to achieve pixel-level vision-language alignment.
In this work, we propose an Acoustically Grounded Cost Learning (AGCL) framework, which transforms the static, audio-agnostic visual-text priors in plain cost learning paradigm~\cite{cho2024cat} into dynamic, audio-grounded cost representations.

\subsection{Cost Volume Learning}
Cost volumes~\cite{rocco2017convolutional,song2021adastereo,liu2022graftnet} were originally introduced to represent dense matching costs between the spatial features of image pairs, measured using metrics such as cosine similarity.
This volume is subsequently aggregated to establish fine-grained correspondences for tasks like optical flow and stereo matching~\cite{cho2022cats++,hong2022cost,hong2022neural,yang2019hierarchical}.
CAT-Seg~\cite{cho2024cat} constructs a multi-modal cost volume between image and text features extracted by CLIP and aggregates it, achieving superior open-vocabulary semantic segmentation performance compared to directly operating on image-text features.
However, existing cost learning paradigms predominantly focus on visual-textual alignments, leaving the potential of acoustic cues unexplored.
In this paper, we further incorporate audio information into the multi-modal cost volume learning process to collaborate with category texts, thereby improving the segmentation of sounding objects from both seen and unseen categories.

\section{Acoustically Grounded Cost Learning}
Given a video clip $\mathcal{V}=\{\mathcal{V}_t\}_{t=1}^T$ of $T$ frames and its corresponding audio clip $\mathcal{A}$, the Open-Vocabulary Audio-Visual Semantic Segmentation (OV-AVSS) task aims to predict frame-level segmentation masks $\bm{Y}=\{\bm{Y}_t\}_{t=1}^T$ of sounding objects over the \textit{total} category set $\mathcal{C}_{\rm{total}} = \mathcal{C}_{\rm{seen}} \cup \mathcal{C}_{\rm{unseen}}$, while only the \textit{seen} categories $\mathcal{C}_{\rm{seen}}$ are annotated and used for training.
The overall architecture of our method is illustrated in Figure~\ref{fig:pipeline}.
The proposed AGCL framework addresses the two primary challenges of the reformulated OV-AVSS task, namely intra-category soundingness discovery and inter-category distractor discrimination.
To tackle the first challenge, we introduce the Audio-Modulated Cost Generation (AMCG) and Audio-Guided Temporal Aggregation (AGTA) modules.
First, the AMCG module softly injects audio features extracted by the audio encoder into the vision encoder of the CLIP model to produce sound-aware visual features.
These features are then correlated with text features to obtain category-wise cost maps that serve as localization priors of sounding objects.
Subsequently, the AGTA module refines these cost embeddings across the temporal dimension to produce temporally consistent representations.
The refined cost embeddings are then fed into a mask decoder to generate pixel-wise segmentation masks for sounding objects in each category.
To tackle the second challenge, we introduce the Synergistic Distractor Mining (SDM) strategy during training.
Rather than learning an optimization bias of universal suppression on most categories, SDM leverages external audio-language knowledge to dynamically mine the most confusing negative categories across both acoustic and semantic dimensions.
By selectively penalizing these distractors, SDM effectively guides the model to learn highly discriminative decision boundaries for better generalization.

\subsection{Multimodal Feature Extraction}
We adopt two pretrained models for feature extraction, namely the CLIP model~\cite{radford2021learning} to extract aligned vision features and category text features, and the CLAP model~\cite{wu2023large} to extract aligned audio features and category text features.
Specifically, the audio clip is first fed into the audio encoder of the CLAP model to extract dense audio features, which are then temporally aligned to the video length $T$, resulting in audio features $\bm{A} \in \mathbb{R}^{T \times D_a}$, where $D_a$ is the channel number.
For the video clip, each frame is passed through the vision encoder of the CLIP model to obtain dense visual features $\bm{V} \in \mathbb{R}^{T \times H \times W \times D_v}$, where $H$, $W$ and $D_v$ denote the height, width, and channel number of visual features.
During this process, the audio features $\bm{A}$ are used as modulation factors to inject audio-aware cues into the visual representations, as detailed in the next section.
For the category names, we obtain two types of text features, \textit{i.e.}, a visual-perspective feature $\bm{E}_v \in \mathbb{R}^{C_{\rm{k}} \times D_{v}}$ from the text encoder of the CLIP model using the template \textit{``A photo of a \{category\}''}, and an acoustic-perspective feature $\bm{E}_a \in \mathbb{R}^{C_{\rm{k}} \times D_{a}}$ from the text encoder of the CLAP model using the template \textit{``The sound of a \{category\}''}.
Unless otherwise specified, in the following sections of this paper, we use $C_{\rm{k}}$ to denote $C_{\rm{seen}} = |\mathcal{C}_{\rm{seen}}|$ during training and $C_{\rm{total}} = |\mathcal{C}_{\rm{total}}|$ during inference.

\subsection{Audio-Modulated Cost Generation}
To obtain audio-aware cost maps with strong generalization, we softly inject audio features into CLIP's vision encoder through audio-conditioned modulation layers, thus discovering sounding regions in each category while preserving the vision-language alignment properties.
Concretely, CLIP's vision encoder consists of $L$ Transformer blocks, where the $l$-th block follows the sequence
$\mathrm{LN}_1^l(\cdot) \rightarrow \mathrm{MHSA}^l(\cdot) \rightarrow \mathrm{LN}_2^l(\cdot) \rightarrow \mathrm{FFN}^l(\cdot)$,
for $l=1,\ldots,L$.
We modify this structure by inserting the audio-conditioned modulation layer after $\mathrm{LN}_2^l(\cdot)$ in selected blocks to modulate the normalized visual activations.
Let $\tilde{\bm{V}}^l \in \mathbb{R}^{T \times H \times W \times D_v}$ denote the visual features output by $\mathrm{LN}_2^l(\cdot)$ in the $l$-th block.
We first apply two linear layers to the audio feature $\bm{A}$ to predict modulation factors:
\begin{equation}
    \bm{\Gamma}^l = 1 + \phi_s^l(\bm{A}), \quad \mathbf{B}^l = \phi_b^l(\bm{A}),
\end{equation}
where $\bm{\Gamma}^l, \mathbf{B}^l \in \mathbb{R}^{T \times D_v}$ are the scale and bias factors, and $\phi_s^l(\cdot)$ and $\phi_b^l(\cdot)$ denote linear layers.
Note that the weights and biases of both $\phi_s^l(\cdot)$ and $\phi_b^l(\cdot)$ are initialized to zero, so that the modulation starts with no effect on visual features and gradually takes effect as training progresses, ensuring a low-intrusive audio injection.
The modulation is then performed as:
\begin{equation}
    \bar{\bm{V}}^l = \bm{\Gamma}^l \odot \tilde{\bm{V}}^l + \mathbf{B}^l,
\end{equation}
where broadcasting is applied along the spatial dimensions, and $\odot$ denotes element-wise multiplication.
The modulated visual features $\bar{\bm{V}}^l$ are subsequently fed into the $\mathrm{FFN}^l(\cdot)$ to produce the output features of the $l$-th block.

After passing through $L$ blocks, the final output $\bm{V}$ of vision encoder becomes audio-aware.
We then calculate the cosine similarity between $\bm{V}$ and the text features $\bm{E}_v$ from CLIP text encoder, to generate audio-aware category-wise cost maps $\mathcal{M}\in \mathbb{R}^{T \times H \times W \times C_{\rm{k}}}$:
\begin{equation}
\label{eq:cost_computation}
    \mathcal{M} = {\rm{Sim}}(\phi^v(\bm{V}), \phi^e(\bm{E}_v)),
\end{equation}
where $\rm{Sim}(\cdot)$ denotes the cosine similarity, and $\phi^v(\cdot)$ and $\phi^e(\cdot)$ are linear projection layers.
We then apply a convolutional layer to the cost map for each category to obtain the cost embeddings
$\bm{M} \in \mathbb{R}^{T \times H \times W \times C_{\rm{k}} \times D_m}$, where $D_m$ is the cost embedding dimension.

\subsection{Multimodal-Guided Cost Refinement}
After obtaining the coarse cost embedding $\bm{M}$, we extend the prior cost-learning paradigm~\cite{cho2024cat} from a single-frame formulation to a temporal one and introduce three aggregation components that progressively refine the embeddings with multimodal cues, thereby improving spatial boundary precision and temporal consistency.

\textbf{Spatial aggregation.}
The spatial aggregation (SA) component refines the category-wise cost embeddings $\bm{M}$ along the spatial dimension to refine object boundary delineation and suppress background noise.
Following~\cite{cho2024cat}, it applies two Swin Transformer blocks~\cite{liu2021swin} to each frame of category-wise cost embedding $\bm{M}(t, :, :, c)\in \mathbb{R}^{H\times W\times D_m}$ for spatial context aggregation, where $t\in[1,T]$, and $c\in[1,C_{\rm{k}}]$.
Visual features from the $t$-th frame $\bm{V}(t)\in \mathbb{R}^{H\times W\times D_v}$ are linearly projected and concatenated with each $\bm{M}(t, :, :, c)$, serving as query-key inputs for the self-attention layers of the Swin Transformer block to provide spatial structural guidance in an attention manner.
Formally, the process can be expressed as:
\begin{equation}
    \bm{M}_{\rm{sa}} = \Psi_{\rm{sa}}(\bm{M},\phi^v_{\rm{sa}}(\bm{V})),
\end{equation}
where $\phi^v_{\rm{sa}}(\cdot)$ denotes the linear projection layer.

\textbf{Audio-guided temporal aggregation.}
Since the input of OV-AVSS consists of multiple video frames, temporal context plays an important role in localizing sound-emitting objects.
Therefore, we further propose an audio-guided temporal aggregation (AGTA) component, which aggregates category-wise cost embeddings across frames based on their audio similarity, thereby enhancing the temporal consistency of the cost representation.
In detail, we apply a Transformer~\cite{vaswani2017attention} layer on each $\bm{M}_{\rm{sa}}(:, h, w, c) \in \mathbb{R}^{T \times D_m}$ to aggregate multi-frame temporal context, where $h \in [1, H], w \in [1, W], c \in [1, C_{\rm{k}}]$.
Formally, the process can be expressed as:
\begin{equation}
    \bm{M}_{\rm{ta}} = \Psi_{\rm{ta}}(\bm{M}_{\rm{sa}},\phi^a_{\rm{ta}}(\bm{A})),
\end{equation}
where the audio feature projected by $\phi^a_{\rm{ta}}(\cdot)$ is utilized as temporal guidance.
Similarly, instead of directly incorporating audio features, AGTA aggregates information through audio-derived cross-frame attention, which helps alleviate overfitting to seen acoustic patterns.

\textbf{Cross-category aggregation.}
We then apply a cross-category aggregation (CCA) component to model the relationships among different categories to enhance category discrimination.
Similar to the previous two modules, the CCA component aggregates cost embeddings among categories for each spatial location using a Transformer layer, guided by category text features $\bm{E}_v$.
The CCA component $\Psi_{\rm{ca}}(\cdot)$ is formulated as follows:
\begin{equation}
    \bm{M}_{\rm{ca}} = \Psi_{\rm{ca}}(\bm{M}_{\rm{ta}},\phi^e_{\rm{ca}}(\bm{E}_v)).
\end{equation}
During training, we also design a category description diversification scheme that leverages an LLM~\cite{achiam2023gpt} to generate multiple semantically equivalent but syntactically diverse descriptions for each category name.
In each iteration, one description is randomly selected to replace the plain category name for text feature extraction as category guidance in the CCA component, thereby enriching the diversity of inter-class interaction patterns.

\begin{table*}[t]
\caption{Comparison with state-of-the-art methods on the AVSBench-OV dataset. Our method obtains significant performance gains for the challenging unseen categories with the same CLIP model as previous methods. * denotes the results are reproduced by us using the authors' released code. Best results are presented in \textbf{bold}.}
\label{tab:sota}
\setlength{\tabcolsep}{8pt}
\centering
\setlength{\arrayrulewidth}{0.6pt}
\resizebox{\linewidth}{!}{
\footnotesize
\begin{tabular}{c|c|cc|cccc}
\hline
\rowcolor[gray]{.9} &  & \textbf{Vision} & \textbf{Text} & \multicolumn{4}{c}{\textbf{mIoU}} \\
\rowcolor[gray]{.9} \multirow{-2}{*}{\textbf{Method}} & \multirow{-2}{*}{\textbf{Reference}} & \textbf{Encoder} & \textbf{Encoder} & \textbf{Unseen} & \textbf{Seen} & \textbf{Harmonic} & \textbf{Overall} \\
\hline\hline
\multicolumn{8}{c}{\textbf{Closed-Set Methods}} \\
\hline
TPAVI~\cite{zhou2022audio,zhou2025audio} & \pub{ECCV'22} & ResNet-50 & - & 0.00 & 28.45  & 0.00 & 18.08 \\
CATR~\cite{li2023catr} & \pub{ACM MM'23} & ResNet-50 & - & 0.00 & 30.64 & 0.00 & 19.73 \\
VCT~\cite{huang2025revisiting}* & \pub{CVPR'25} & Swin-Base & - & 0.00 & 52.09 & 0.00 & 32.00 \\
\hline\hline
\multicolumn{8}{c}{\textbf{Zero-Shot Method}} \\
\hline
Sam4AVS~\cite{yu2023can} & \pub{BMVC'23} & Grounded-SAM & - & 8.53 & 13.55 & 10.47 & 12.47 \\
\hline\hline
\multicolumn{8}{c}{\textbf{Open-Vocabulary Methods}} \\
\hline
OV2Seg~\cite{wang2023towards} & \pub{ICCV'23} & ResNet-50 & CLIP & 12.49 & 36.25  & 18.58 & 26.93 \\
OpenVIS~\cite{guo2025openvis} & \pub{AAAI'25} & ResNet-50 & CLIP-ViT-B/32 & 17.54 & 45.23 & 25.31 & 34.40 \\
CLIP-VIS~\cite{zhu2024clip} & \pub{TCSVT'24} & ResNet-50 & CLIP & 15.00 & 40.35 & 21.87 & 30.31 \\
\multirow{2}{*}{OV-AVSS~\cite{guo2024open}} & \multirow{2}{*}{\pub{ACM MM'24}} & R-50+CLIP-B & CLIP-ViT-B/16 & 22.00 & 49.77  & 30.71 & 38.91 \\
& & Swin-B+CLIP-L & CLIP-ViT-L/14 & 29.14 & 55.43  & 38.20 & 44.81 \\
\rowcolor{blue!8} & & CLIP-ViT-B/16 & CLIP-ViT-B/16 & 33.81& 61.03 & 43.51 & 50.69 \\
\rowcolor{blue!8} \multirow{-2}{*}{\textbf{AGCL (Ours)}} & \multirow{-2}{*}{-} & CLIP-ViT-L/14 & CLIP-ViT-L/14 & \textbf{45.59} & \textbf{61.47} & \textbf{52.35} & \textbf{55.20} \\
\hline
\end{tabular}}
\end{table*}

\begin{figure}[t]
   \centering
   \includegraphics[width=\linewidth]{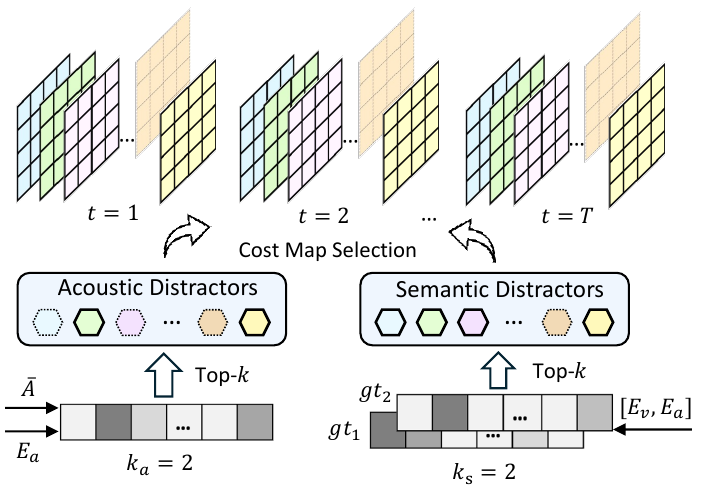}
   \caption{A simplified example of Synergistic Distractor Mining. $k_a$ acoustic distractors are selected based on the feature similarities between the input audio and category texts. $k_s$ semantic distractors are selected based on text feature similarities between ground-truth and other categories. Here, $k_a$ and $k_s$ are set to $2$ for better illustration.}
   \label{fig:hncm}
\end{figure}

\subsection{Synergistic Distractor Mining}
To tackle the challenge of inter-category distractor discrimination, we propose a Synergistic Distractor Mining (SDM) strategy, which selectively mines acoustic and semantic distractor categories, to sharpen the decision boundaries during training.
We present a simplified example in Figure~\ref{fig:hncm} for illustration.
For acoustic distractor mining, we first calculate the similarity between CLAP audio features $\bm{A}$ and text features $\bm{E}_a$ as:
\begin{equation}
\bm{R}_a = {\rm{Sim}}(\phi^a_{\rm{sim}}(\bar{\bm{A}}), \phi^e_{\rm{sim}}(\bm{E}_a)),
\end{equation}
where $\bar{\bm{A}} \in \mathbb{R}^{1 \times D_a}$ is the temporal average of $\bm{A}$, $\phi^a_{\rm{sim}}(\cdot)$ and $\phi^e_{\rm{sim}}(\cdot)$ are projection layers.
The top-$k_a$ categories with the highest similarities in $\bm{R}_a$ are regarded as acoustic distractors, whose indices are denoted as $\mathcal{I}_a$.
To make the category selection process differentiable, we apply Gumbel-TopK~\cite{kool2019stochastic, jang2016categorical, maddison2016concrete} to $\bm{R}_a$.
A soft audio-category distribution is first generated as:
\begin{equation}
    \tilde{\bm{R}}_a = {\rm{Softmax}}(\bm{R}_a + \bm{G}), \bm{G} \sim \rm{Gumbel}(0, 1).
\end{equation}
We then define a multi-row one-hot selection matrix $\mathbf{S}^{\text{hard}}_a \in \mathbb{R}^{k_a\times C_{\rm{seen}}}$ as follows:
\begin{equation}
\mathbf{S}^{\text{hard}}_a(i,j) =
\begin{cases}
1, & \text{if } j = \mathcal{I}_a(i), \\
0, & \text{otherwise.}
\end{cases}
\end{equation}
To keep gradients to $\bm{R}_a$, we replicate $\tilde{\bm{R}}_a$ into $\bm{R}^{'}_a \in \mathbb{R}^{k_a \times C_{\rm{seen}}}$ and apply the straight-through estimator:
\begin{equation}
    \bm{S}_a = \bm{S}^{\rm{hard}}_a + \bm{R}^{'}_a - {\rm{sg}}(\bm{R}^{'}_a)
\end{equation}
where ${\rm{sg}}(\cdot)$ stops gradients.
The cost maps for acoustic distractors can be obtained by matrix multiplication between $\mathcal{M}$ and $\bm{S}_a$.

For semantic distractor mining, we compute the pairwise similarity matrix $\bm{R}_s \in \mathbb{R}^{C_{\rm{seen}} \times C_{\rm{seen}}}$ among the seen categories based on the concatenation of their CLIP text features $\bm{E}_v$ and CLAP text features $\bm{E}_a$.
For each ground-truth category, the top-$k_s$ categories having the highest similarities to it in $\bm{R}_s$ are selected as semantic distractors, as they are semantically similar to the true sounding category but belong to different categories.
The cost maps for semantic distractors are obtained similarly to those for acoustic distractors.
Note that the SDM strategy is applied only during training, while all categories are included for cost computing during inference.

\subsection{Loss Functions}
During training, the refined cost embedding $\bm{M}_{ca}$ is fed into a mask decoder~\cite{cho2024cat} to predict segmentation logits $\bm{Y}$.
We compute BCE loss $\mathcal{L}_{\rm{mask}}$ between $\bm{Y}$ and the ground-truth mask $\bm{Y}^*$ to supervise the semantic segmentation learning on seen categories.
To ensure semantic consistency across modalities, we further design an alignment loss between the CLIP- and CLAP-encoded text features, encouraging implicit alignment between visual and audio representations.
The similarity matrix $\bm{P} \in \mathbb{R}^{C_{\rm{seen}} \times C_{\rm{seen}}}$ is computed as:
\begin{equation}
\bm{P} = {\rm{Sim}}(\phi^e_{\rm{clip}}(\bm{E}_v), {\phi^e_{\rm{clap}}(\bm{E}_a)}),
\end{equation}
where $\phi^e_{\rm{clip}}(\cdot)$ and $\phi^e_{\rm{clap}}(\cdot)$ are projection layers.
The ground-truth for feature similarity is an identity matrix $\bm{P}^*$.
The alignment loss $\mathcal{L}_{\rm{align}}$ is also implemented as a BCE loss between $\bm{P}$ and $\bm{P}^*$.
$\mathcal{L}_{\rm{align}}$ encourages features from the same category to be close while pushing apart those from different categories, thereby preserving semantic consistency across modalities.
The overall loss function of our model is summarized as:
\begin{equation}
    \mathcal{L} = \lambda_{mask}\mathcal{L}_{\rm{mask}} + \lambda_{align}\mathcal{L}_{\rm{align}},
\end{equation}
where $\lambda_{mask}$ and $\lambda_{align}$ are the loss coefficients.

\section{Experiment}
\subsection{Dataset and Evaluation Metrics}
We train and evaluate our models on the AVSBench-OV dataset~\cite{guo2024open}, which is derived from AVSBench-Semantic~\cite{zhou2025audio}.
AVSBench-OV contains $70$ semantic categories in total, of which $40$ frequently occurring categories are used as seen categories, and the other $30$ uncommon ones are reserved as unseen categories.
Videos containing the unseen categories are excluded from the training set to avoid data leakage.
The dataset includes $5{,}184$ real-world videos for training, $1{,}240$ for validation, and $1{,}490$ for testing.
Following the prior work~\cite{guo2024open}, we adopt mean Intersection-over-Union (mIoU) as the evaluation metric.
Four variants are reported, namely the overall mIoU over all $70$ categories, Seen-mIoU for seen categories, Unseen-mIoU for unseen categories, and Harmonic-mIoU defined as the harmonic mean between Seen- and Unseen-mIoUs.
We also conduct out-of-distribution (OOD) evaluation on the VPO dataset~\cite{chen2024unraveling} to validate the cross-dataset generalization capability of our method.

\subsection{Implementation Details}
Our model is implemented using PyTorch~\cite{pytorch}.
We employ CLIP-ViT-B/16 and CLIP-ViT-L/14 models~\cite{radford2021learning} as the pre-trained vision-language models for fair comparison with the previous method~\cite{guo2024open}, and adopt the CLAP model~\cite{wu2023large} pre-trained on the LAION-Audio-630K dataset as the audio-language model.
During training, only the key and value projection layers are finetuned for both CLIP and CLAP models.
For each video clip, the total number of video frames $T$ is set to $5$ during training, while we sample $1$ frame per second during testing.
The input video frame resolution is $336{\times}336$, and the cost embedding dimension $D_m$ is $128$.
The numbers of selected categories in SDM are $k_a = 10$ and $k_s = 3$.
The loss coefficients $\lambda_{mask}$ and $\lambda_{align}$ are set to $1.0$ and $0.01$.
The AdamW optimizer is employed with an initial learning rate of $1e^{-4}$ for CLIP-ViT-B/16 and $2e^{-4}$ for CLIP-ViT-L/14 with a weight decay of $1e^{-4}$ and a cosine learning-rate schedule.
We train our model for a total of $40{,}000$ iterations with a batch size of $1$.
All experiments are conducted on a single RTX 4090 GPU with $24$ GB of memory.

\subsection{Comparison with State-of-the-art Methods}
Since OV-AVSS is a relatively new task, there has been only one dedicated work~\cite{guo2024open} prior to ours. 
Following its evaluation protocol, we report results on the AVSBench-OV dataset using the same set of adapted baselines, which include closed-set AVS methods~\cite{zhou2022audio, li2023catr, huang2025revisiting}, a zero-shot AVS method~\cite{yu2023can}, and open-vocabulary semantic segmentation methods~\cite{wang2023towards, guo2025openvis, zhu2024clip}.
As shown in Table~\ref{tab:sota}, our AGCL framework achieves significant improvements over all prior methods across both seen and unseen categories.
In particular, AGCL (CLIP-ViT-B/16) achieves a harmonic mIoU of $43.51\%$, exceeding the previous best OV-AVSS (CLIP-ViT-L/14) by nearly $5\%$ on the unseen set.
When using the same CLIP model (\textit{i.e.}, CLIP-ViT-B/16), our method outperforms OV-AVSS by over $10\%$ on both seen and unseen categories.
These improvements indicate that our AGCL framework learns more stable and transferable representations of sounding objects.
Furthermore, when equipped with a stronger CLIP model, AGCL delivers an additional performance gain, reaching $45.59\%$ unseen mIoU and $61.47\%$ seen mIoU.

\subsection{Ablation Studies and Further Discussion}
We conduct ablation studies on the AVSBench-OV dataset with CLIP-ViT-B/16 as the basic model to verify the effectiveness of different designs in our method.

\begin{table}[!htbp]
\caption{Ablation study on different components of our AGCL framework. \textit{SDM-A} and \textit{SDM-S} denote mining acoustically and semantically confusing distractors, respectively.}
\label{tab:ablation_component}
\setlength{\tabcolsep}{2pt}
\centering
\setlength{\arrayrulewidth}{0.4pt}
\resizebox{\linewidth}{!}{
\tiny
\begin{tabular}{cc|cc|cccc}
\hline
\rowcolor[gray]{.9} & & & & \multicolumn{4}{c}{\textbf{mIoU}} \\
\rowcolor[gray]{.9} \multirow{-2}{*}{\textbf{AMCG}} & \multirow{-2}{*}{\textbf{AGTA}} & \multirow{-2}{*}{\textbf{SDM-A}} & \multirow{-2}{*}{\textbf{SDM-S}} & \textbf{Unseen} & \textbf{Seen} & \textbf{Harmonic} & \textbf{Overall} \\
\hline\hline
& & & & 26.47 & 56.66 & 36.08 & 44.40 \\
\hline
\checkmark & & & & 29.11 & 58.66 & 38.91 & 46.91 \\
& \checkmark & & & 28.44 & 58.29 & 38.23 & 46.67 \\
\checkmark & \checkmark & & & 30.45 & 60.65 & 40.54 & 48.87 \\
\hline
\checkmark & \checkmark & \checkmark & &  32.32 & 59.27 & 41.83 & 48.58 \\
\checkmark & \checkmark &  & \checkmark & 31.81 & 60.30 & 41.65 & 48.72 \\
\checkmark & \checkmark & \checkmark & \checkmark  & \textbf{33.81} & \textbf{61.03} & \textbf{43.51} & \textbf{50.69} \\
\hline
\end{tabular}}
\end{table}

\textbf{Component Analysis.}
We report the ablation study results of each proposed components of our AGCL framework in Table~\ref{tab:ablation_component}.
The first row represents the baseline cost-learning paradigm without audio injection or distractor mining.
In the second row, generating sound-aware cost maps with AMCG brings notable improvements on both seen and unseen categories.
In the third row, incorporating the AGTA component in the cost-refinement phase further enhances temporal coherence by propagating sounding-region features across frames. Although its gain is slightly smaller than AMCG due to audio being injected at a later phase, the improvement in harmonic mIoU confirms that temporal consistency benefits overall segmentation stability.
When combining AMCG and AGTA in the fourth row, the model achieves an overall gain of nearly +$4.5\%$ mIoU compared to the baseline, showing the superiority of our low-intrusive audio-injection mechanism, which discovers sounding objects without disturbing the vision-language alignment.

We further analyze the design of the synergistic distractor mining strategy in the last three rows. 
When only acoustically confusing distractors or only semantically confusing distractors are mined, the model achieves higher Unseen-mIoU, suggesting improved generalization to unseen categories.
However, their performance on seen categories drops, possibly because limiting the negative scope reduces the diversity of seen category supervision and weakens intra-category compactness.
By jointly considering both acoustically and semantically confusing distractors, the full AGCL framework increases the performance on seen categories and further improves unseen category segmentation, yielding the best harmonic mIoU of $43.51\%$.
This indicates that combining complementary types of distractors exposes the model to more discriminative boundaries, thus improving open-vocabulary generalization.

\begin{table}[!htbp]
\caption{Ablation study on different approaches of incorporating audio cues for cost generation in our AMCG module.}
\label{tab:ablation_inject}
\centering
\setlength{\arrayrulewidth}{0.4pt}
\resizebox{\linewidth}{!}{
\tiny
\begin{tabular}{c|cccc}
\hline
\rowcolor[gray]{.9} & \multicolumn{4}{c}{\textbf{mIoU}} \\
\rowcolor[gray]{.9} \multirow{-2}{*}{\textbf{Method}}& \textbf{Unseen} & \textbf{Seen} & \textbf{Harmonic} & \textbf{Overall} \\
\hline\hline
 - & 26.47 & 56.66 & 36.08 & 44.40 \\
SA & 26.68 & 57.78 & 36.51 & 45.40 \\
CA & 27.43 & \textbf{58.92} & 37.44 & 46.10 \\
\textbf{AMCG} & \textbf{29.11} & 58.66 & \textbf{38.91} & \textbf{46.91} \\
\hline
\end{tabular}}
\end{table}

\begin{figure*}[!t]
    \centering
    \includegraphics[width=\linewidth]{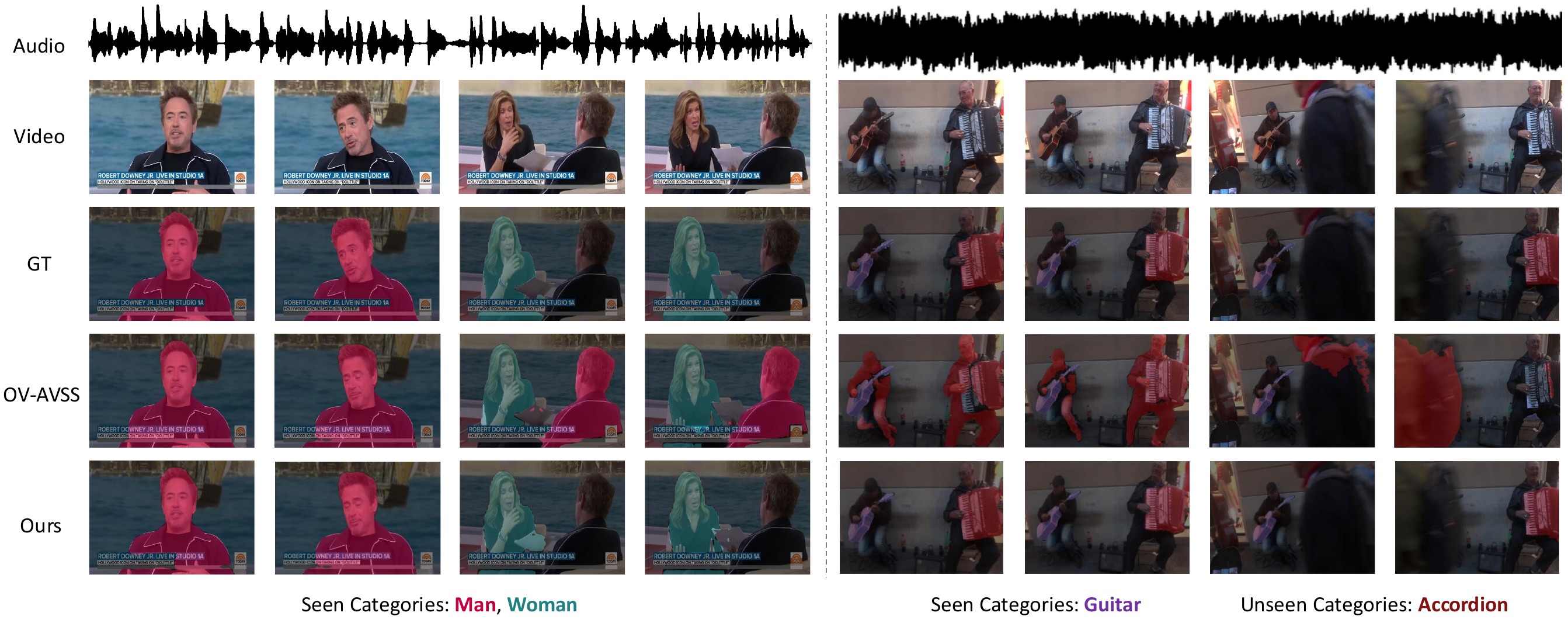}
    \caption{Qualitative comparison between OV-AVSS~\cite{guo2024open} and our AGCL on both seen and unseen categories. Sounding categories correspond to masks of the same colors.}
    \label{fig:seg_results}
\end{figure*}

\begin{figure}[!htbp]
    \centering
    \includegraphics[width=\linewidth]{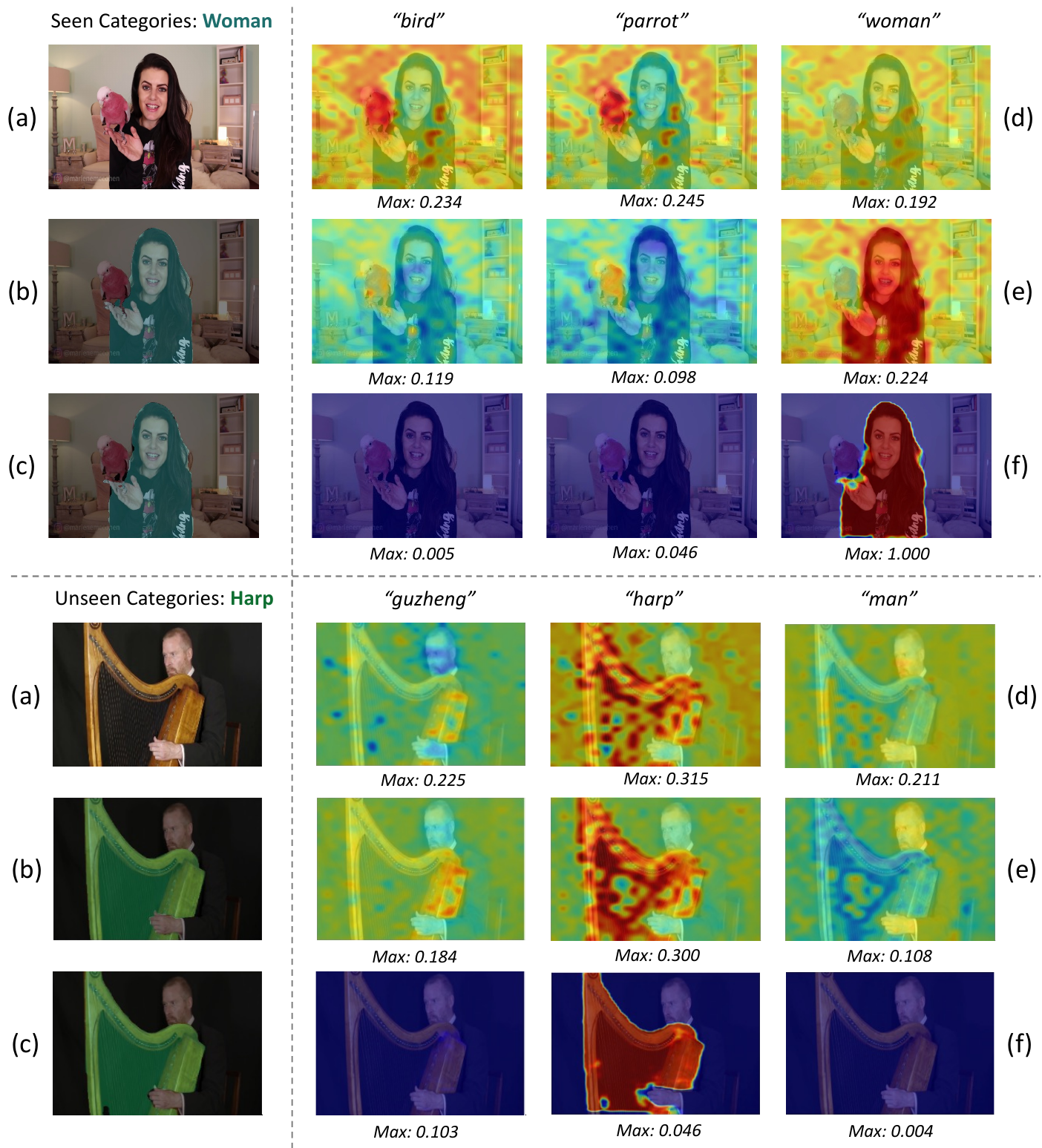}
    \caption{Visualization of cost maps. (a) Original video frame. (b) Ground-truth mask. (c) Our prediction. (d) Cost maps generated by the original CLIP model. (e) Cost maps generated in our AMCG module. (f) Cost maps after refinement.}
    \label{fig:cost}
\end{figure}

\textbf{Audio Cues Incorporation.}
We further compare different approaches for incorporating audio cues in our AMCG module in Table~\ref{tab:ablation_inject}.
To make visual features audio-aware while retaining vision-language alignment, we explore three low-intrusive designs: (1) SA, which concatenates audio and visual tokens in the self-attention layers, (2) CA, which injects audio as keys and values in cross-attention adapters, and (3) AMCG, which generates modulation factors with audio features.
All inserted modules are zero-initialized to avoid disturbing the pretrained representation at the early training iterations.
Among the three, AMCG achieves the best performance, particularly on unseen categories (+$2.64\%$ \textit{vs.} +$0.96\%$ or +$0.21\%$), indicating stronger open-vocabulary generalization.
Compared with SA and CA, which directly alter the feature composition of CLIP representations, AMCG performs soft feature rescaling that highlights sounding regions while maintaining the original distribution.

\begin{table}[!htbp]
\caption{Ablation study on different types of guidance features used in the AGTA module.}
\label{tab:ablation_temp}
\centering
\setlength{\arrayrulewidth}{0.4pt}
\resizebox{\linewidth}{!}{
\tiny
\begin{tabular}{c|cccc}
\hline
\rowcolor[gray]{.9} & \multicolumn{4}{c}{\textbf{mIoU}} \\
\rowcolor[gray]{.9} \multirow{-2}{*}{\textbf{Guidance}}& \textbf{Unseen} & \textbf{Seen} & \textbf{Harmonic} & \textbf{Overall} \\
\hline\hline
\textit{w/o} AGTA & 29.11 & 58.66 & 38.91 & 46.91 \\
No Guide & 29.27 & 60.00 & 39.34 & 48.00 \\
Visual & 29.70 & 60.01 & 39.73 & 48.19 \\
\textbf{Audio} & \textbf{30.45} & \textbf{60.65} & \textbf{40.54} & \textbf{48.87} \\
\hline
\end{tabular}}
\end{table}

\textbf{Temporal Guidance Feature.}
As shown in Table~\ref{tab:ablation_temp}, we ablate the types of guidance features used in the AGTA module.
Compared with the baseline without temporal aggregation, using either no guidance (the second row) or visual features to guide attention (the third row) improves performance mainly on seen categories, as both variants enhance the temporal consistency of cost embeddings.
However, their gain on unseen categories remains marginal.
In contrast, using audio features as guidance yields the best results, especially for unseen categories, as the audio-based cross-frame similarity serves as a robust indicator of sounding continuity and enables more precise temporal aggregation.

\begin{table}[!htbp]
\caption{Evaluation results of our AGCL \textit{w/o} training on the VPO dataset. ``OOD'' and ``ID'' denotes out-of-distribution and in-distribution, respectively. \textbf{Best} and \underline{Second Best}.}
\label{tab:ood_vpo}
\centering
\setlength{\arrayrulewidth}{0.8pt}
\resizebox{\linewidth}{!}{
\begin{tabular}{c|c|ccc|ccc}
\hline
\rowcolor[gray]{.9} & & \multicolumn{3}{c}{\textbf{VPO (MS)}} & \multicolumn{3}{c}{\textbf{VPO (MSMI)}} \\
\rowcolor[gray]{.9} \multirow{-2}{*}{\textbf{Method}} & \multirow{-2}{*}{\textbf{Type}} & \textbf{MIoU$\uparrow$} & \textbf{$F_\beta \uparrow$} & \textbf{FDR $\downarrow$} & \textbf{MIoU$\uparrow$} & \textbf{$F_\beta \uparrow$} & \textbf{FDR $\downarrow$} \\
\hline\hline
TPAVI~\cite{zhou2022audio} & ID & 44.08 & 58.14 & 30.82 & 50.37 & 66.80 & 29.82 \\
AVSegFormer~\cite{gao2024avsegformer} & ID & 56.46 & 71.89 & 24.66 & 50.96 & 64.96 & 32.72 \\
CAVP~\cite{chen2024unraveling} & ID & \textbf{61.85} & \textbf{74.60} & \textbf{20.24} & \textbf{57.22} & \textbf{72.26} & \textbf{24.04} \\
\hline
OV-AVSS~\cite{guo2024open} & OOD & 45.04 & 56.42 & 43.51 & 42.30 & 53.18 & 47.59 \\
\textbf{AGCL (Ours)} & OOD & \underline{57.72} & \underline{72.91} & \underline{24.05} & \underline{54.24} & \underline{68.65} & \underline{29.52} \\
\hline
\end{tabular}}
\end{table}

\textbf{Out-of-Distribution Evaluation.}
As shown in Table~\ref{tab:ood_vpo}, we evaluate the generalization capability of our method on the VPO dataset~\cite{chen2024unraveling} using mono audios under an Out-of-Distribution (OOD) setting.
Specifically, OV-AVSS~\cite{guo2024open} and our AGCL are trained solely on the AVSBench-OV dataset using CLIP-ViT-L/14 as backbone and directly evaluated on the VPO dataset.
The results indicate that our method significantly outperforms the OOD baseline OV-AVSS across both MS and MSMI subsets. 
Furthermore, our approach even surpasses TPAVI~\cite{zhou2022audio} and AVSegFormer~\cite{gao2024avsegformer}, which are evaluated under an In-Distribution (ID) setting with training on the VPO dataset, achieving the second-best performance overall.
Outperforming an in-distribution-trained model without target-domain fine-tuning demonstrates our method's robust generalization.

\subsection{Qualitative Analysis}
Figure~\ref{fig:seg_results} compares our method with OV-AVSS~\cite{guo2024open}.
For seen categories such as \textit{man} and \textit{woman}, OV-AVSS fails to identify the active speaker when the speaker changes, whereas our method accurately follows these dynamic transitions.
This benefits from AGTA, which aggregates temporal context using audio similarities, and AMCG, which enhances sounding regions while suppressing silent but visually salient areas.
For the unseen \textit{accordion} category, OV-AVSS produces over-expanded masks that include surrounding people and incompletely cover the instrument, while our method segments the full instrument with clear boundaries.
Figure~\ref{fig:cost} visualizes cost maps for seen and unseen categories.
After AMCG, visually present but non-sounding categories are substantially suppressed, demonstrating that audio-conditioned modulation highlights sounding regions and attenuates visual distractors.
In the top row, the maximum responses of ``\textit{bird}'' and ``\textit{parrot}'' decrease to about half of those from the original CLIP model.
In the bottom row, the model distinguishes the sounding ``\textit{harp}'' from the acoustically similar ``\textit{guzheng}'' and the silent ``\textit{man}''.
Subsequent multimodal-guided refinement further suppresses non-sounding entities and concentrates responses on ``\textit{woman}'' and ``\textit{harp}'', leading to accurate predictions.

\section{Conclusion}
Open-Vocabulary Audio-Visual Semantic Segmentation (OV-AVSS) segments sound-emitting objects from both seen and unseen categories.
We reformulate its foreground definition from generic soundingness to category-specific audio-visual patterns and propose Acoustically Grounded Cost Learning (AGCL) to transform static, audio-agnostic visual-text priors into dynamic, audio-grounded cost representations.
Our designed Audio-Modulated Cost Generation (AMCG) and Audio-Guided Temporal Aggregation (AGTA) modules enable low-intrusive audio injection for intra-category soundingness discovery, while our introduced Synergistic Distractor Mining (SDM) strategy mines acoustically and semantically confusing negatives for inter-category discrimination.
Extensive experiments on the AVSBench-OV dataset show substantial improvements of our AGCL over previous methods.

\begin{acks}
This research was supported in part by the National Natural Science Foundation of China (Grant No. 62502142, 62572166, 62302140, 62472139), the Natural Science Foundation of Anhui Province (Grant No. 2508085QF226), and the Fundamental Research Funds for the Central Universities (Grant No. JZ2025HGTA0161).
The computation is completed on the HPC Platform of Hefei University of Technology.
\end{acks}

\bibliographystyle{ACM-Reference-Format}
\balance
\bibliography{sample-base}

@inproceedings{zhou2022audio,
  title={Audio--visual segmentation},
  author={Zhou, Jinxing and Wang, Jianyuan and Zhang, Jiayi and Sun, Weixuan and Zhang, Jing and Birchfield, Stan and Guo, Dan and Kong, Lingpeng and Wang, Meng and Zhong, Yiran},
  booktitle={ECCV},
  year={2022}
}

@article{zhou2025audio,
  title={Audio-visual segmentation with semantics},
  author={Zhou, Jinxing and Shen, Xuyang and Wang, Jianyuan and Zhang, Jiayi and Sun, Weixuan and Zhang, Jing and Birchfield, Stan and Guo, Dan and Kong, Lingpeng and Wang, Meng and others},
  journal={IJCV},
  year={2025}
}

@inproceedings{huang2023discovering,
  title={Discovering sounding objects by audio queries for audio visual segmentation},
  author={Huang, Shaofei and Li, Han and Wang, Yuqing and Zhu, Hongji and Dai, Jiao and Han, Jizhong and Rong, Wenge and Liu, Si},
  booktitle={IJCAI},
  year={2023}
}

@inproceedings{li2023catr,
  title={Catr: Combinatorial-dependence audio-queried transformer for audio-visual video segmentation},
  author={Li, Kexin and Yang, Zongxin and Chen, Lei and Yang, Yi and Xiao, Jun},
  booktitle={ACM MM},
  year={2023}
}

@inproceedings{wang2024prompting,
  title={Prompting segmentation with sound is generalizable audio-visual source localizer},
  author={Wang, Yaoting and Liu, Weisong and Li, Guangyao and Ding, Jian and Hu, Di and Li, Xi},
  booktitle={AAAI},
  year={2024}
}

@inproceedings{gao2024avsegformer,
  title={Avsegformer: Audio-visual segmentation with transformer},
  author={Gao, Shengyi and Chen, Zhe and Chen, Guo and Wang, Wenhai and Lu, Tong},
  booktitle={AAAI},
  year={2024}
}

@inproceedings{yang2024cooperation,
  title={Cooperation does matter: Exploring multi-order bilateral relations for audio-visual segmentation},
  author={Yang, Qi and Nie, Xing and Li, Tong and Gao, Pengfei and Guo, Ying and Zhen, Cheng and Yan, Pengfei and Xiang, Shiming},
  booktitle={CVPR},
  year={2024}
}

@inproceedings{li2024selm,
  title={Selm: Selective mechanism based audio-visual segmentation},
  author={Li, Jiaxu and Yu, Songsong and Wang, Yifan and Wang, Lijun and Lu, Huchuan},
  booktitle={ACM MM},
  year={2024}
}

@inproceedings{huang2025revisiting,
  title={Revisiting Audio-Visual Segmentation with Vision-Centric Transformer},
  author={Huang, Shaofei and Ling, Rui and Hui, Tianrui and Li, Hongyu and Zhou, Xu and Zhang, Shifeng and Liu, Si and Hong, Richang and Wang, Meng},
  booktitle={CVPR},
  year={2025}
}

@inproceedings{liu2024audio,
  title={Audio-visual segmentation via unlabeled frame exploitation},
  author={Liu, Jinxiang and Liu, Yikun and Zhang, Fei and Ju, Chen and Zhang, Ya and Wang, Yanfeng},
  booktitle={CVPR},
  year={2024}
}

@inproceedings{zha2025implicit,
  title={Implicit Counterfactual Learning for Audio-Visual Segmentation},
  author={Zha, Mingfeng and Li, Tianyu and Wang, Guoqing and Wang, Peng and Wu, Yangyang and Yang, Yang and Shen, Heng Tao},
  booktitle={ICCV},
  year={2025}
}

@inproceedings{guo2024open,
  title={Open-vocabulary audio-visual semantic segmentation},
  author={Guo, Ruohao and Qu, Liao and Niu, Dantong and Qi, Yanyu and Yue, Wenzhen and Shi, Ji and Xing, Bowei and Ying, Xianghua},
  booktitle={ACM MM},
  year={2024}
}

@inproceedings{ding2022decoupling,
  title={Decoupling zero-shot semantic segmentation},
  author={Ding, Jian and Xue, Nan and Xia, Gui-Song and Dai, Dengxin},
  booktitle={CVPR},
  year={2022}
}

@article{ding2022open,
  title={Open-vocabulary universal image segmentation with maskclip},
  author={Ding, Zheng and Wang, Jieke and Tu, Zhuowen},
  journal={arXiv preprint arXiv:2208.08984},
  year={2022}
}

@inproceedings{huynh2022open,
  title={Open-vocabulary instance segmentation via robust cross-modal pseudo-labeling},
  author={Huynh, Dat and Kuen, Jason and Lin, Zhe and Gu, Jiuxiang and Elhamifar, Ehsan},
  booktitle={CVPR},
  year={2022}
}

@inproceedings{liang2023open,
  title={Open-vocabulary semantic segmentation with mask-adapted clip},
  author={Liang, Feng and Wu, Bichen and Dai, Xiaoliang and Li, Kunpeng and Zhao, Yinan and Zhang, Hang and Zhang, Peizhao and Vajda, Peter and Marculescu, Diana},
  booktitle={CVPR},
  year={2023}
}

@inproceedings{xu2022simple,
  title={A simple baseline for open-vocabulary semantic segmentation with pre-trained vision-language model},
  author={Xu, Mengde and Zhang, Zheng and Wei, Fangyun and Lin, Yutong and Cao, Yue and Hu, Han and Bai, Xiang},
  booktitle={ECCV},
  year={2022}
}

@inproceedings{liu2024open,
  title={Open-vocabulary segmentation with semantic-assisted calibration},
  author={Liu, Yong and Bai, Sule and Li, Guanbin and Wang, Yitong and Tang, Yansong},
  booktitle={CVPR},
  year={2024}
}

@inproceedings{li2025mask,
  title={Mask-Adapter: The Devil is in the Masks for Open-Vocabulary Segmentation},
  author={Li, Yongkang and Cheng, Tianheng and Feng, Bin and Liu, Wenyu and Wang, Xinggang},
  booktitle={CVPR},
  year={2025}
}

@inproceedings{xu2022groupvit,
  title={Groupvit: Semantic segmentation emerges from text supervision},
  author={Xu, Jiarui and De Mello, Shalini and Liu, Sifei and Byeon, Wonmin and Breuel, Thomas and Kautz, Jan and Wang, Xiaolong},
  booktitle={CVPR},
  year={2022}
}

@inproceedings{xu2023open,
  title={Open-vocabulary panoptic segmentation with text-to-image diffusion models},
  author={Xu, Jiarui and Liu, Sifei and Vahdat, Arash and Byeon, Wonmin and Wang, Xiaolong and De Mello, Shalini},
  booktitle={CVPR},
  year={2023}
}

@inproceedings{xu2023side,
  title={Side adapter network for open-vocabulary semantic segmentation},
  author={Xu, Mengde and Zhang, Zheng and Wei, Fangyun and Hu, Han and Bai, Xiang},
  booktitle={CVPR},
  year={2023}
}

@inproceedings{cho2024cat,
  title={Cat-seg: Cost aggregation for open-vocabulary semantic segmentation},
  author={Cho, Seokju and Shin, Heeseong and Hong, Sunghwan and Arnab, Anurag and Seo, Paul Hongsuck and Kim, Seungryong},
  booktitle={CVPR},
  year={2024}
}

@inproceedings{xie2024sed,
  title={Sed: A simple encoder-decoder for open-vocabulary semantic segmentation},
  author={Xie, Bin and Cao, Jiale and Xie, Jin and Khan, Fahad Shahbaz and Pang, Yanwei},
  booktitle={CVPR},
  year={2024}
}

@inproceedings{zhao2025dpseg,
  title={DPSeg: Dual-Prompt Cost Volume Learning for Open-Vocabulary Semantic Segmentation},
  author={Zhao, Ziyu and Li, Xiaoguang and Shi, Lingjia and Imanpour, Nasrin and Wang, Song},
  booktitle={CVPR},
  year={2025}
}

@inproceedings{rocco2017convolutional,
  title={Convolutional neural network architecture for geometric matching},
  author={Rocco, Ignacio and Arandjelovic, Relja and Sivic, Josef},
  booktitle={CVPR},
  year={2017}
}

@inproceedings{liu2022graftnet,
  title={Graftnet: Towards domain generalized stereo matching with a broad-spectrum and task-oriented feature},
  author={Liu, Biyang and Yu, Huimin and Qi, Guodong},
  booktitle={CVPR},
  year={2022}
}

@inproceedings{song2021adastereo,
  title={Adastereo: A simple and efficient approach for adaptive stereo matching},
  author={Song, Xiao and Yang, Guorun and Zhu, Xinge and Zhou, Hui and Wang, Zhe and Shi, Jianping},
  booktitle={CVPR},
  year={2021}
}

@article{cho2022cats++,
  title={Cats++: Boosting cost aggregation with convolutions and transformers},
  author={Cho, Seokju and Hong, Sunghwan and Kim, Seungryong},
  journal={TPAMI},
  year={2022}
}

@inproceedings{hong2022cost,
  title={Cost aggregation with 4d convolutional swin transformer for few-shot segmentation},
  author={Hong, Sunghwan and Cho, Seokju and Nam, Jisu and Lin, Stephen and Kim, Seungryong},
  booktitle={ECCV},
  year={2022}
}

@article{hong2022neural,
  title={Neural matching fields: Implicit representation of matching fields for visual correspondence},
  author={Hong, Sunghwan and Nam, Jisu and Cho, Seokju and Hong, Susung and Jeon, Sangryul and Min, Dongbo and Kim, Seungryong},
  journal={NeurIPS},
  year={2022}
}

@inproceedings{yang2019hierarchical,
  title={Hierarchical deep stereo matching on high-resolution images},
  author={Yang, Gengshan and Manela, Joshua and Happold, Michael and Ramanan, Deva},
  booktitle={CVPR},
  year={2019}
}

@inproceedings{radford2021learning,
  title={Learning transferable visual models from natural language supervision},
  author={Radford, Alec and Kim, Jong Wook and Hallacy, Chris and Ramesh, Aditya and Goh, Gabriel and Agarwal, Sandhini and Sastry, Girish and Askell, Amanda and Mishkin, Pamela and Clark, Jack and others},
  booktitle={ICML},
  year={2021}
}

@inproceedings{wu2023large,
  title={Large-scale contrastive language-audio pretraining with feature fusion and keyword-to-caption augmentation},
  author={Wu, Yusong and Chen, Ke and Zhang, Tianyu and Hui, Yuchen and Berg-Kirkpatrick, Taylor and Dubnov, Shlomo},
  booktitle={ICASSP},
  year={2023}
}

@inproceedings{liu2021swin,
  title={Swin transformer: Hierarchical vision transformer using shifted windows},
  author={Liu, Ze and Lin, Yutong and Cao, Yue and Hu, Han and Wei, Yixuan and Zhang, Zheng and Lin, Stephen and Guo, Baining},
  booktitle={ICCV},
  year={2021}
}

@article{vaswani2017attention,
  title={Attention is all you need},
  author={Vaswani, Ashish and Shazeer, Noam and Parmar, Niki and Uszkoreit, Jakob and Jones, Llion and Gomez, Aidan N and Kaiser, {\L}ukasz and Polosukhin, Illia},
  journal={NeurIPS},
  year={2017}
}

@article{jang2016categorical,
  title={Categorical reparameterization with gumbel-softmax},
  author={Jang, Eric and Gu, Shixiang and Poole, Ben},
  journal={arXiv preprint arXiv:1611.01144},
  year={2016}
}

@article{maddison2016concrete,
  title={The concrete distribution: A continuous relaxation of discrete random variables},
  author={Maddison, Chris J and Mnih, Andriy and Teh, Yee Whye},
  journal={arXiv preprint arXiv:1611.00712},
  year={2016}
}

@article{pytorch,
  author  = {Adam Paszke and Sam Gross and Francisco Massa and Adam Lerer and James Bradbury and Gregory Chanan and Trevor Killeen and Zeming Lin and Alban Desmaison and Andreas Kopf and Edward Gibson and Chien Yu Wang and Zachary DeVito and Martin Auer and Nikita Nikitin and Sergiy Popov and Jonathan Dynan and Colby Childers and Arnaud Lefebvre and Soumith Chintala},
  title   = {PyTorch: An Imperative Style, High-Performance Deep Learning Library},
  journal = {Journal of Machine Learning Research},
  year    = {2024}
}

@inproceedings{wang2023towards,
  title={Towards open-vocabulary video instance segmentation},
  author={Wang, Haochen and Yan, Cilin and Wang, Shuai and Jiang, Xiaolong and Tang, Xu and Hu, Yao and Xie, Weidi and Gavves, Efstratios},
  booktitle={ICCV},
  year={2023}
}

@inproceedings{guo2025openvis,
  title={OpenVIS: Open-vocabulary video instance segmentation},
  author={Guo, Pinxue and Huang, Hao and He, Peiyang and Liu, Xuefeng and Xiao, Tianjun and Zhang, Wenqiang},
  booktitle={AAAI},
  year={2025}
}

@article{zhu2024clip,
  title={CLIP-VIS: Adapting CLIP for open-vocabulary video instance segmentation},
  author={Zhu, Wenqi and Cao, Jiale and Xie, Jin and Yang, Shuangming and Pang, Yanwei},
  journal={TCSVT},
  year={2024},
}

@inproceedings{yu2023can,
  title={How Can Contrastive Pre-training Benefit Audio-Visual Segmentation? A Study from Supervised and Zero-shot Perspectives.},
  author={Yu, Jiarui and Li, Haoran and Hao, Yanbin and Wu, Jinmeng and Xu, Tong and Wang, Shuo and He, Xiangnan},
  booktitle={BMVC},
  year={2023}
}

@article{achiam2023gpt,
  title={Gpt-4 technical report},
  author={Achiam, Josh and Adler, Steven and Agarwal, Sandhini and Ahmad, Lama and Akkaya, Ilge and Aleman, Florencia Leoni and Almeida, Diogo and Altenschmidt, Janko and Altman, Sam and Anadkat, Shyamal and others},
  journal={arXiv preprint arXiv:2303.08774},
  year={2023}
}

@inproceedings{kool2019stochastic,
  title={Stochastic beams and where to find them: The gumbel-top-k trick for sampling sequences without replacement},
  author={Kool, Wouter and Van Hoof, Herke and Welling, Max},
  booktitle={ICML},
  year={2019}
}

@inproceedings{chen2024unraveling,
  title={Unraveling instance associations: A closer look for audio-visual segmentation},
  author={Chen, Yuanhong and Liu, Yuyuan and Wang, Hu and Liu, Fengbei and Wang, Chong and Frazer, Helen and Carneiro, Gustavo},
  booktitle={CVPR},
  year={2024}
}

@inproceedings{wang2024ref,
  title={Ref-avs: Refer and segment objects in audio-visual scenes},
  author={Wang, Yaoting and Sun, Peiwen and Zhou, Dongzhan and Li, Guangyao and Zhang, Honggang and Hu, Di},
  booktitle={ECCV},
  year={2024}
}

@inproceedings{sun2024unveiling,
  title={Unveiling and Mitigating Bias in Audio Visual Segmentation},
  author={Sun, Peiwen and Zhang, Honggang and Hu, Di},
  booktitle={ACM MM},
  year={2024}
}

@inproceedings{liu2023audio,
  title={Audio-visual segmentation by exploring cross-modal mutual semantics},
  author={Liu, Chen and Li, Peike Patrick and Qi, Xingqun and Zhang, Hu and Li, Lincheng and Wang, Dadong and Yu, Xin},
  booktitle={ACM MM},
  year={2023}
}

@inproceedings{di2021video,
  title={Video background music generation with controllable music transformer},
  author={Di, Shangzhe and Jiang, Zeren and Liu, Si and Wang, Zhaokai and Zhu, Leyan and He, Zexin and Liu, Hongming and Yan, Shuicheng},
  booktitle={ACM MM},
  year={2021}
}

@inproceedings{zhang2025moma,
  title={Moma-kitchen: A 100k+ benchmark for affordance-grounded last-mile navigation in mobile manipulation},
  author={Zhang, Pingrui and Gao, Xianqiang and Wu, Yuhan and Liu, Kehui and Wang, Dong and Wang, Zhigang and Zhao, Bin and Ding, Yan and Li, Xuelong},
  booktitle={ICCV},
  year={2025}
}

@article{ju2026instruction,
  title={From Instruction to Event: Sound-Triggered Mobile Manipulation},
  author={Ju, Hao and Huang, Shaofei and Li, Hongyu and Ding, Zihan and Liu, Si and Wang, Meng and Zheng, Zhedong},
  journal={arXiv preprint arXiv:2601.21667},
  year={2026}
}

@inproceedings{he2024progressive,
  title={Progressive feature self-reinforcement for weakly supervised semantic segmentation},
  author={He, Jingxuan and Cheng, Lechao and Fang, Chaowei and Feng, Zunlei and Mu, Tingting and Song, Mingli},
  booktitle={AAAI},
  year={2024}
}

@inproceedings{wu2024masked,
  title={Masked collaborative contrast for weakly supervised semantic segmentation},
  author={Wu, Fangwen and He, Jingxuan and Yin, Yufei and Hao, Yanbin and Huang, Gang and Cheng, Lechao},
  booktitle={WACV},
  year={2024}
}

@inproceedings{he2024customize,
  title={Customize your nerf: Adaptive source driven 3d scene editing via local-global iterative training},
  author={He, Runze and Huang, Shaofei and Nie, Xuecheng and Hui, Tianrui and Liu, Luoqi and Dai, Jiao and Han, Jizhong and Li, Guanbin and Liu, Si},
  booktitle={CVPR},
  year={2024}
}

@inproceedings{huang2025unleashing,
  title={Unleashing the temporal-spatial reasoning capacity of gpt for training-free audio and language referenced video object segmentation},
  author={Huang, Shaofei and Ling, Rui and Li, Hongyu and Hui, Tianrui and Tang, Zongheng and Wei, Xiaoming and Han, Jizhong and Liu, Si},
  booktitle={AAAI},
  year={2025}
}

@inproceedings{wang2023transferring,
  title={Transferring CLIP's knowledge into zero-shot point cloud semantic segmentation},
  author={Wang, Yuanbin and Huang, Shaofei and Gao, Yulu and Wang, Zhen and Wang, Rui and Sheng, Kehua and Zhang, Bo and Liu, Si},
  booktitle={ACM MM},
  year={2023}
}

\end{document}